# Pattern Inversion as a Pattern Recognition Method for Machine Learning

Alexei Mikhailov[*], Mikhail Karavay

Institute for Control Problems, Russian Academy of Sciences, Moscow, Russia
alxmikh@gmail.com, mkaravay@yandex.ru

**Abstract:** Artificial neural networks use a lot of coefficients that take a great deal of computing power for their adjustment, especially if deep learning networks are employed. However, there exist coefficients-free extremely fast indexing-based technologies that work, for instance, in Google search engines, in genome sequencing, etc. The paper discusses the use of indexing-based methods for pattern recognition. It is shown that for pattern recognition applications such indexing methods replace with inverse patterns the fully inverted files, which are typically employed in search engines. Not only such inversion provide automatic feature extraction, which is a distinguishing mark of deep learning, but, unlike deep learning, pattern inversion supports almost instantaneous learning, which is a consequence of absence of coefficients. The paper discusses a pattern inversion formalism that makes use on a novel pattern transform and its application for unsupervised instant learning. Examples demonstrate a view-angle independent recognition of three-dimensional objects, such as cars, against arbitrary background, prediction of remaining useful life of aircraft engines, and other applications. In conclusion, it is noted that, in neurophysiology, the function of the neocortical mini-column has been widely debated since 1957. This paper hypothesize that, mathematically, the cortical mini-column can be described as an inverse pattern, which physically serves as a connection multiplier expanding associations of inputs with relevant pattern classes.

## 1. Introduction

Deep learning is characterized by a high computational complexity of training, because of a large number of neural nodes' coefficients that need to be calculated. At the same time, the inverse patterns make it feasible to avoid using any coefficients at all, which opens up the possibility of instant learning, which boils down to establishing connections between features and classes of patterns. When recognizing image objects, the objects' pixels are used as features at the lowest level. As a result, with instant learning, as with deep learning, there is no need to engineer the features manually. Instead, each pixel is considered as a 3-dimensional pattern containing red, green and blue components, or as a multidimensional pattern, which additionally includes infrared, ultraviolet and other components.

This paper provides a definition of inverse patterns [1], which is closely related to the definition of inverse sets [2, 3]. This concept is related to reverse file technology that is used in search engines [4], as well as in the word-in-a-bag method, where images of objects are searched for in video clips [5]. However, in the latter case, the visual objects are encoded in the form of texts, whereas with inverse patterns, the objects are encoded with numerical features.

Inverse patterns are obtained by a transformation of original patterns, - the transformation that is discussed in Section 3. Whereas a pattern is traditionally specified by a feature vector or a feature set, an inverse pattern is represented by a set of pattern classes. The pattern transformation connects features with classes. An example of such connections is shown in Figure 1, where the pattern {*b, g*}, which is given as a pair of features, is classified using the links shown on the left side of the Figure 1. Mathematically, the pattern class can be found by intersecting the inverse patterns (see the columnar diagram in the right-hand side of Figure. 1).

---

[*] Corresponding author

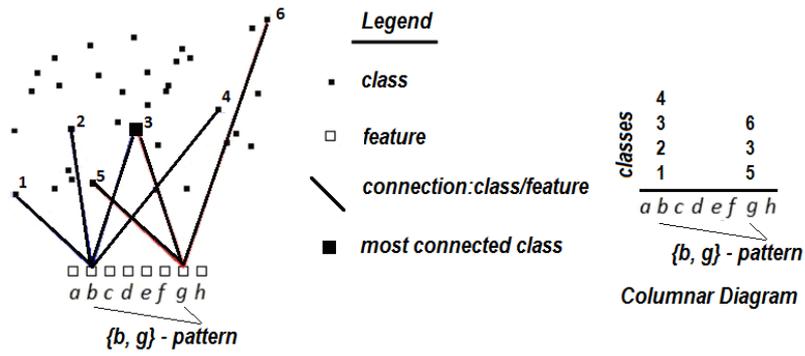

*Figure 1. Connectivity diagram.*

Connections of features with classes are shown. The pattern {*b*, *g*} should be assigned to class 3, since this class is associated with most features, in this case, with two. For comparison, class 4 is associated with a single feature, which is *b*.

The paper discusses results of an object recognition experiment, where shots of 3D-objects (cars) were made from different view angles against an arbitrary background. The results are presented in section 2.

The formalism of inverse patterns-based recognition method is discussed in section 3. Inverse patterns-based prediction of real useful life of aircraft engines is considered in section 3.6. Comments are provided in section 4.

## 2. Results

The examples below show cars against an arbitrary background at view angles that were used for training (Figure 2).

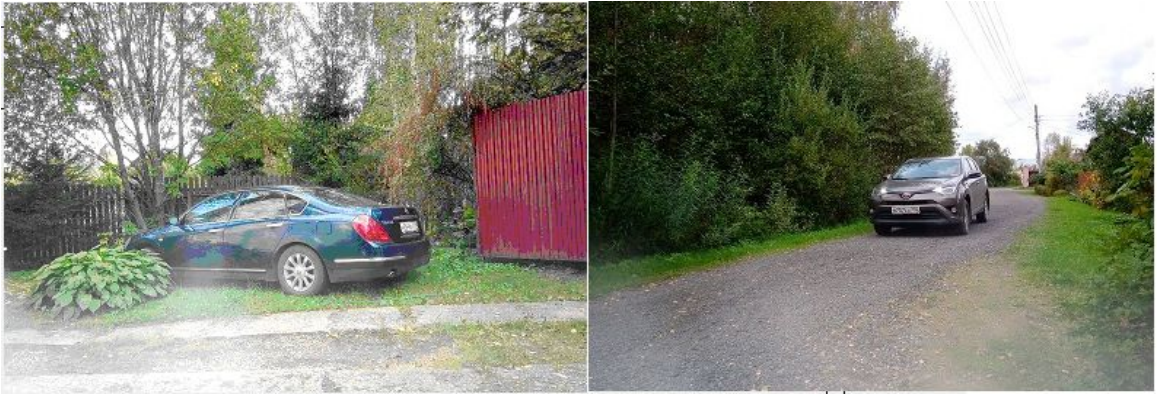

*Figure 2. Two car models at specific view angles used for training*

At testing, invariance of object recognition with respect to view angles was demonstrated. The right-hand side of Figure 3 shows the first model at a testing angle that is different from training angle, whereas the left-hand side shows the model retrieved from the database.

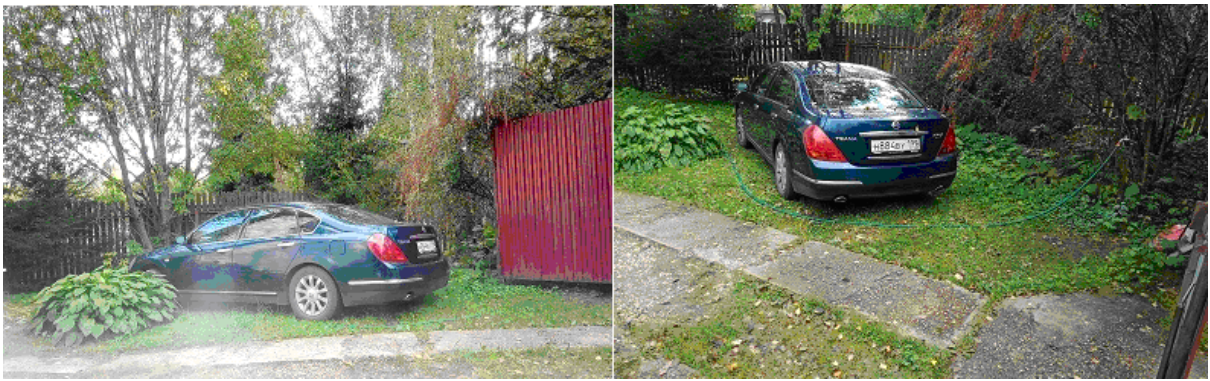

*Figure 3. Training and testing view angles of the first model.*

A pair of histograms that are rotated 180° relative to each other for ease of comparison is presented in Figure 4. The horizontal axis represents pixel classes, whereas frequencies of these classes are put on vertical axes $H_1$, $H_2$.

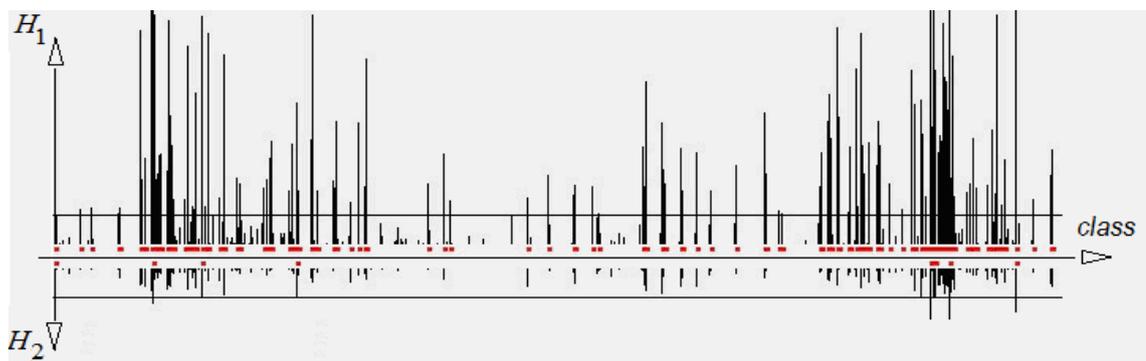

*Figure 4. Signature histograms of the first model.*

The upper histogram represents the left-hand side car (Figure 3), and the lower histogram represents the right-hand side car (Figure 3). There are 8 matching pairs of samples among the red dots - the samples, whose heights exceed thresholds that are marked with horizontal lines.
It can be seen that histograms of the original and rotated cars remain almost symmetrical about the horizontal axis (not taking into account the amplitudes of classes). A similar result for the second car model can be seen from Figures 5, 6.

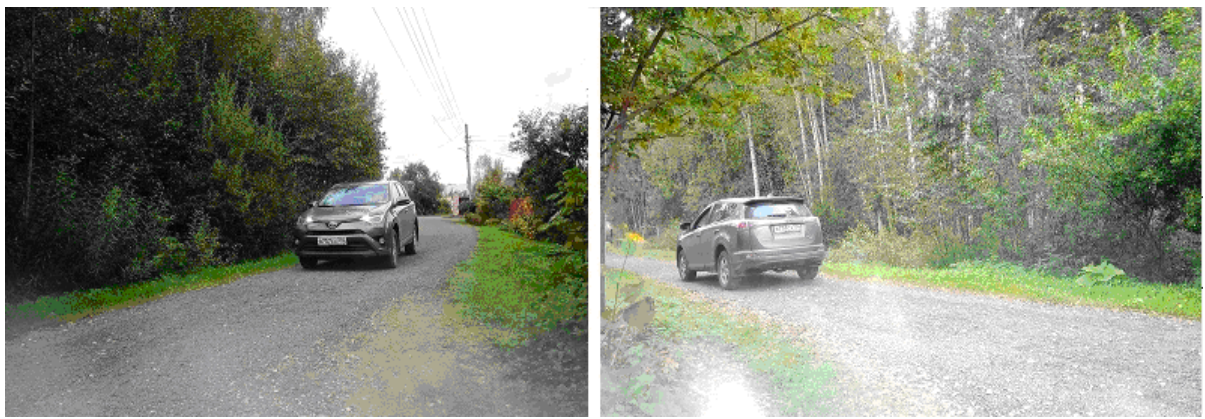

*Figure 5. Training and testing view angles of the second model.*

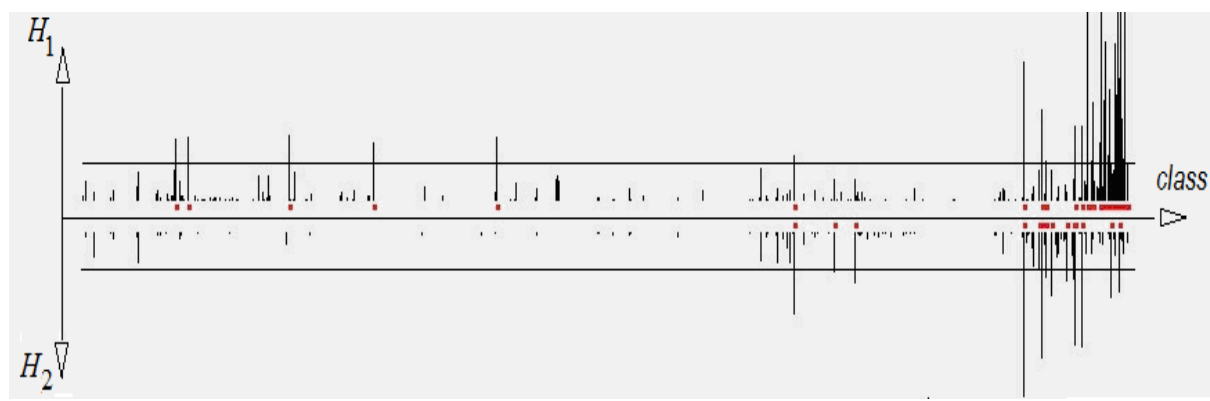

*Figure 6. Signature histograms of the second model.*

Here, the upper histogram represents the left-hand side car (Figure 5), and the lower histogram represents the right-hand side car (Figure 5). There are 8 matching pairs of samples among the red dots - the samples, whose heights exceed thresholds that are marked with horizontal lines.

The symmetry is broken when histograms of different models (Figure 7) are compared.

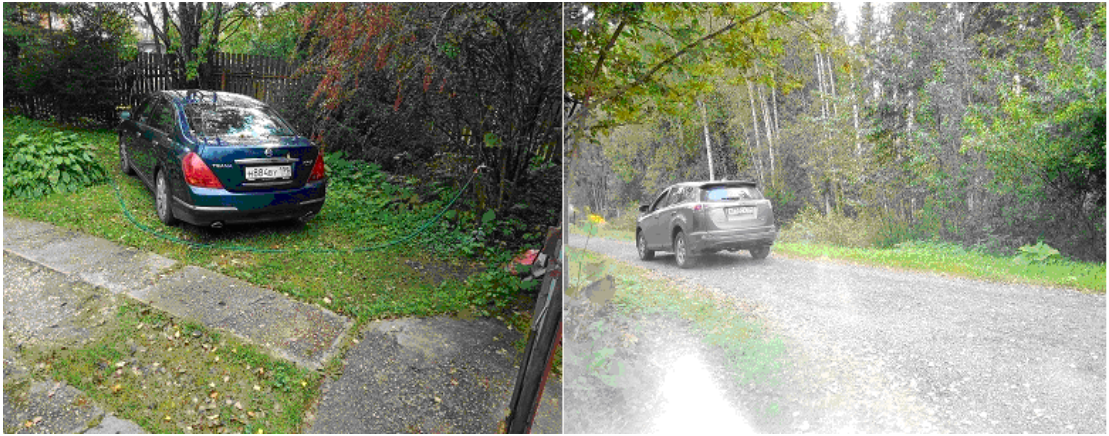

*Figure 7. First and second model at testing view-angles*

The histograms of different models were found to have no common features (Figure 8). Note that positions of samples on the horizontal axis are not related to distances between classes, so that if two samples are close neighbors, the classes they represent may be quite dissimilar.
The symmetry of histograms underlies the invariance of recognition with respect to the view angle and, within certain bounds, the size of the object.

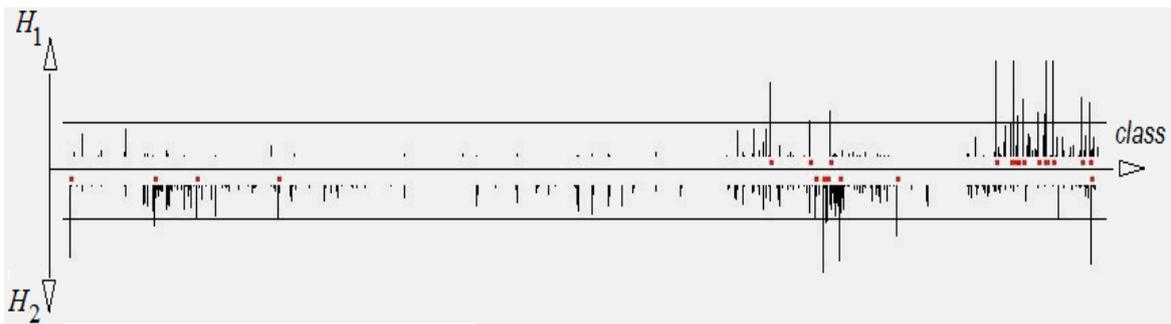

*Figure 8. Asymmetric signature histograms of different models.*

## 3. Pattern inversion formalism

### *3.1. Introduction to pattern recognition by pattern inversion*

For this method, classification of a pattern is achieved by finding a distribution of classes contained in inverse patterns. The position of a maximum of the distribution determines the most probable class. The distribution is obtained by counting the connections between features and classes contained in the inverse patterns. In Figure. 9, where input pattern features are encircled, the inverse patterns are marked with double circles that contain classes associated with these features.

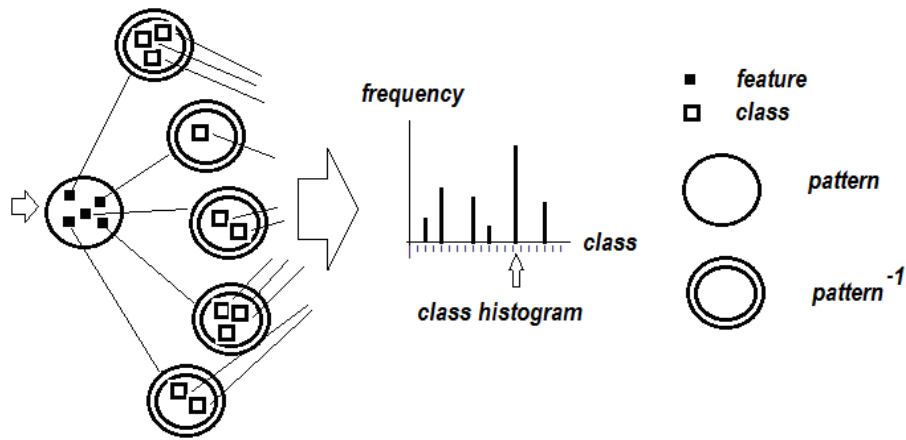

*Figure 9. Diagram of inverse patterns-based pattern recognition method.*

### 3.2. Specifics of multilevel classification

The recognition scheme (Figure 9) illustrates the operation of a single level of the system. More levels may be needed when a sequence of object images is used as an input. Then a sequence of classes of recognized objects can be considered as a sequence of meta-features, and a class of this sequence, once it is identified at the second level, becomes a meta-class. Recognition at each level is performed using inverse patterns of corresponding levels. Note that the set of meta-features at the input of each level is represented in the form of a histogram, whose samples are fed into next level.

### 3.3. Specifics of training

All levels work in the same way. At each level, the pattern is classified using previously created inverse patterns containing the internal classes of this level. If classification fails, the input pattern is assigned to a newly created internal class that updates the previously created inverse patterns. Such procedure amounts to an unsupervised learning. Depending on the requirements of the problem being solved, a teacher can be introduced by adding a table that assigns to each inner class an external class. The number of emerging inner classes is controlled by a generalization radius (section 3.7). Usually, all features of the first level are normalized to a common integer range, which practically does not affect the classification accuracy, which is proportional to the number of emerging internal classes. At the same time, the generalizing ability of the classification system is inversely proportional to the number of inner classes.

At each level, the classes, which are output by a previous level, serve as features, whereas a histogram of these classes serves as an input pattern. For example, when recognizing a sequence of moving objects a three level structure can be used. At the first level, the pattern is a pixel, that is, its color components. At the second level, the pattern is a histogram of pixel classes obtained as a result of recognizing a cluster of object pixels. At the third level, the pattern is a histogram of object classes. Then the output of the third level is the class of the sequence or the type of movement, for example, the meaning of the sequence of gestures, the name of the dance, the nature of the object's evolution (the transformation of a tadpole into a frog), etc.

### 3.4. Specifics of training and recognition when working with image objects shown against an arbitrary background.

The topic of highlighting objects against arbitrary background is not directly related to the topic of inverse patterns. But, since such application was considered in experiments, the used approach is briefly described below. On training, a moving object is to be highlighted in a difference image, that is, in an image that is a difference of two consecutive frames. Alternatively, to obtain a difference image the image without an object is subtracted from the image with an object, for example, a road without a car is subtracted from a road with a car. When comparing two images, a threshold sliding window was used as a filter. After that, the system is trained on the difference image. Finally, for verification, the image without an object is recognized. If the latter image is mistaken for an object, then pixel classes found in no-object image are masked, that is, the pixels classes, whose frequency exceeds a certain threshold. After that, the training procedure moves on to the next object.

On recognition, the entire image (with or without an object) is scanned. All pixels belonging to the unmasked classes are selected and grouped into clusters using any clustering algorithm, for instance, a propagating wave that paints connected pixels, - the pixels that are located within a chosen distance from each other. Finally, the pixel clusters are sequentially recognized at the second level that produces a distribution of cluster activities, the total maximum of which determines the class of the winning object.

*3.5. Formalism of inverse patterns-based classification method*

Let all variables be integers and the number $N$ of pattern classes be represented by $K$-dimensional feature vectors

$$\mathbf{x}_n = (x_{n,1}, x_{n,2}, ..., x_{n,k}, ..., x_{n,K}), \quad n = 1, 2, ..., N \tag{1}$$

Here the distance between any two classes $p$ and $q$ is greater than a given number $R$, that is, $|\mathbf{x}_p - \mathbf{x}_q| \geq R$; the variable $x_{n,k}$ represents the value of the ($0 \leq x_{n,k} < X$) $k$-th feature of the $n$-th vector.

Problem: For a given vector $\mathbf{x} = \{x(k), k = 1, 2, ..., K\}$, find a class $n$, such that $|\mathbf{x} - \mathbf{x}_n| < R$. If the class $n$ does not exist, expand the list (1) with the vector $\mathbf{x}_{N+1} = \mathbf{x}$.

This problem can be solved with inverse patterns, which are defined as class sets indexed by the features

$$\{n\}_{x,k} = \{\forall n : |x_{n,k} - x| < R\}, \quad k = 1, 2, ..., K, \quad x = 0, 1, ..., X-1 \tag{2}$$

Figure 10 illustrates the notion of inverse patterns that are depicted as columns of dots, where each dot represents a class.

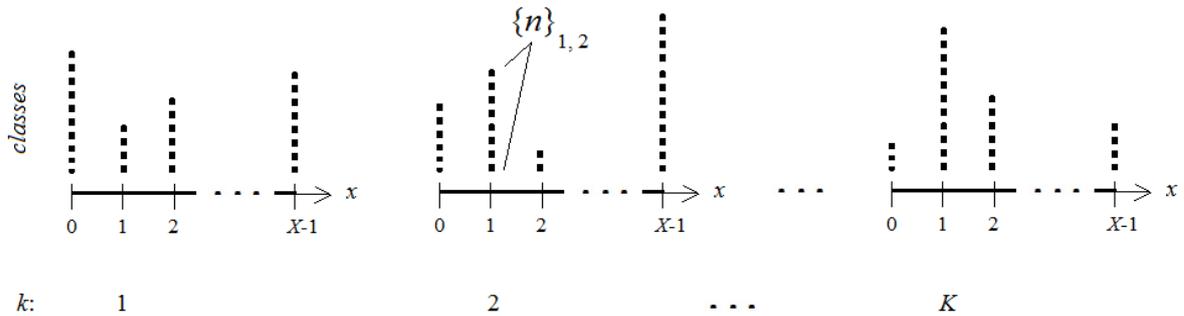

*Figure 10. Inverse patterns divided into K groups of columns*

In case of numeric features, $x(k), k = 1, 2, ..., K$, a histogram $H_R(n|\mathbf{x})$ of classes contained in inverse patterns can be obtained with the following algorithm:

$$\begin{array}{c} r = 0, \pm 1, ..., \pm R, \quad k = 1, 2, ..., K, \\ \forall n \in \{n\}_{x(k)+r, k}, \quad H_R(n|\mathbf{x})++ \end{array} \tag{3}$$

The maximum of the class histogram is equal to the dimensionality $K$ of the patterns, and position of the maximum determines the most active class (winner). This statement is consequence of the following properties of the pattern transform (2).

Each dimension, $k = 1, 2, ..., K$, of the transform (2)

- contains $N$ different classes: $\sum_{x=0}^{X-1} |\{n\}_{x,k}| = N$
- does not contain intersecting sets: $\{n\}_{x,k} \cap \{n\}_{y,k} = \varnothing$

The categorical features appear as a binary code after a threshold processing of output samples of previous levels. For categorical features the transform (2) can be re-written as

$$\{n\}_{x,k} = \{\forall n : x_{n,k} = x\}, \quad k = 1, 2, ..., K, \quad x = 0, 1$$

Here $K$ is the number of all categories and $x_{n,k} \in \{0,1\}$.

For a categorical pattern $x(k)$, $k = 1, 2, ..., K$, ( $x(k) \in \{0,1\}$ ) a categorical class histogram $H(n|\mathbf{x})$ is obtained with the algorithm that is similar to (3)

$$k = 1, 2, ..., K, \forall n \in \{n\}_{x,k}, H(n|\mathbf{x}) + +$$

The maximum of the categorical class histogram is not pre-determined by dimensionality as it is not available in case of patterns represented by feature sets. But, similar to the numerical case, the position of its maximum determines the winner.

On training, if the maximum $H_{max}$ of the class histogram is less than $K$ (numerical case) or less then a pre-set recognition threshold (categorical case), the number of classes has to be incremented as $N = N+1$, and a new class introduced into the corresponding inverse patterns

$$\{n\}_{x(k), k} = \{n\}_{x(k), k} \cup N, k = 1, 2, ..., K$$

### 3.6. Prediction problem

For prediction problems, each vector $n$ from (1) is associated with a certain parameter $t_n$, for instance, with the time. Besides, no distance restrictions are imposed on vectors (1). Next, the sets are introduced that contain the values of the predicted parameter $t_n$, which were observed together with the values of the features $x_{n,k}$

$$\{t_n\}_{x,k} = \{t_n : x_{n,k} = x\}, k = 1, 2, ..., K, x = 0, 1, ..., X-1 \quad (4)$$

Upon arrival of a feature vector $x(k)$, $k = 1, 2, ..., K$, the histogram of the predicted parameter $t$ is calculated as

$$k = 1, 2, ..., K, \forall t \in \{t\}_{x(k), k}, H(t|\mathbf{x}) + + \quad (5)$$

The position of its maximum determines the most probable value of the predicted parameter. In [3], the data from [6] were used, which show the remaining useful life of 100 aircraft engines, the engine parameters in 20731 flights (training data) and the engine parameters in another 13196 flights (test data). Each flight is represented by 26 - dimensional vector. Examples of histograms are shown in Figure 11.

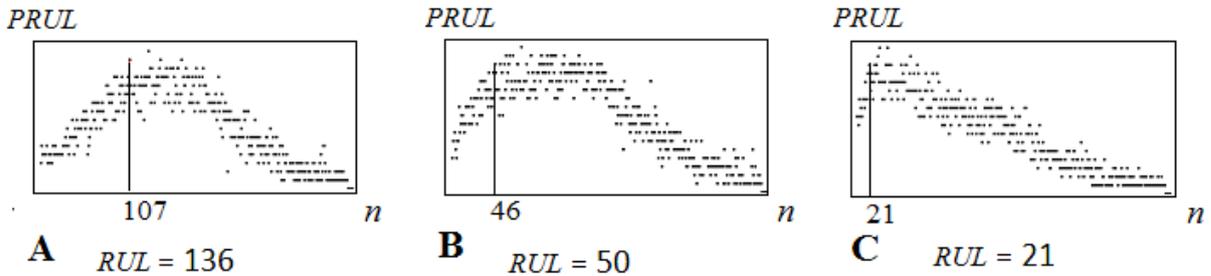

*Figure 11. Probability of engine failure as a function of the operating cycle n (flight).*

Here, PRUL stands for the probable real useful life and RUL stands for the real useful life. The prediction errors are as follows: (**A**) PRUL / RUL = 107/136. (**B**) PRUL / RUL = 46/50. (**C**) PRUL / RUL = 21/21.

It is interesting to note that as the time of failure approaches, the distribution symmetry breaks down. Taking this observation into account makes it possible to increase the reliability of the prediction, ensuring the absence of critical errors when the predicted lifetime exceeds the actual remaining lifetime. For comparison, other methods [7-10] for engine RUL prediction are punctuated by critical errors.

The training on a laptop (1.2 GHz) for all 20731 flights took 0.5 sec and the prediction for 13196 flights took 0.8 sec [3].

### 3.7. Generalization radius

The generalization radius ( $0 \leq R < X$ ), alias a receptive field is set as a percentage of the feature range $X$. The influence of the radius $R$ is illustrated in Figure 12, where the pixels of the leftmost image [11] should be assigned to one of 3 external classes: water, buildings and vegetation. For this, each pixel is treated as a 3-dimensional pattern with red, green and blue components. The training is carried out by a teacher who links the first external class to automatically found inner classes of pixels from a sub-area of water, the second external class is linked to the inner classes from a sub-area of buildings, and the third external class is linked to the inner classes from a sub-area of vegetation. The sub-areas used for training are enclosed by white ovals. The created look-up table is needed to classify all pixels outside the sub-areas. At $R = 0$, the number of inner classes $N$ approaches the number of image points. Since the size of sub-areas is much smaller than that of the image, the table contains much fewer internal classes, so that most pixels remain unclassified (middle picture, Figure 12). However, with $R = 10\%$ of the feature range 0 - 255, almost all pixels outside the sub-areas fall into the same classes as pixels in the training areas, though the number of internal classes is small ($N = 13$). Therefore, almost all points of the image were marked. The result is shown in the rightmost image, Figure 12.

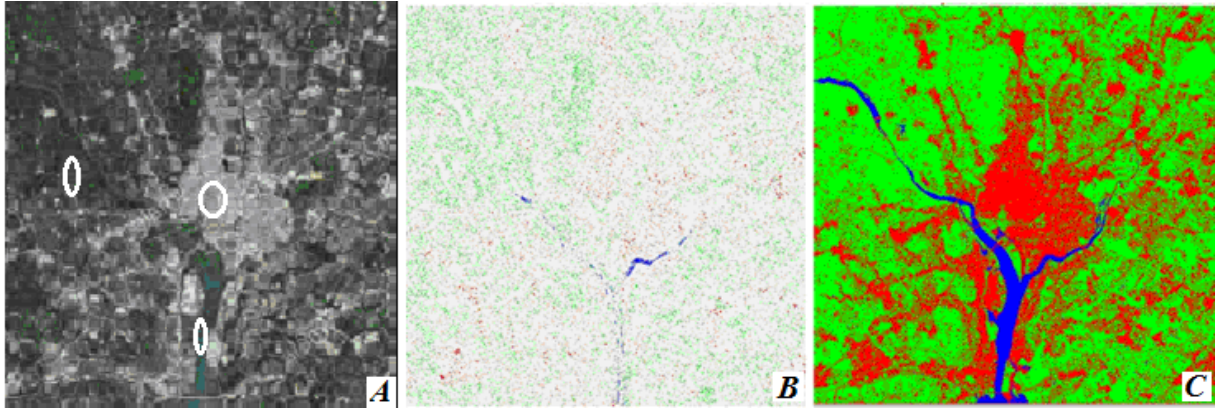

*Figure 12. Image segmentation.*

In Figure 12, ***A*** is the original image to be segmented, where training sub-areas are marked by white curves, ***B*** is the image that shows a result of low quality segmentation at $R = 0$, and the image ***C*** shows a high quality segmentation with a better generalization ($R = 10\%$).

*3.8. Computational complexity*

The computational complexity $C$ of classifying a $K$-dimensional pattern with algorithm (2) is proportional to the average height $h$ of inverse patterns: $C \sim Kh$, where $h = \dfrac{1}{|S|} \sum_{(x,k) \in S} |\{n\}_{x,k}|$, $S = \{(x,k): \ |\{n\}_{x,k}| > 0\}$. The value of $h$ can be controlled as it is proportional to the number of inner classes $N$ that decreases with increasing radius $R$ and range $X$. There is a fast version of algorithm (2), which calculates a histogram of only those classes $n$ at the step $k$, for which $H_R(n\,|\,\mathbf{x}) = k$. For very large problems with hundreds of thousands and millions of inner classes $N$, this number can keep on increasing as long as the average height $h$ remains low and, accordingly, the training time remains small. Once this boundary is reached, new classes should be placed in a second algorithmically identical module working in parallel, and so on.

## 4. Discussion

The software-algorithmic simplicity of expressions (2) - (3) for recognition systems and expressions (4) - (5) for prediction systems makes it possible to build these systems as micro-devices on local platforms, for instance, flexible programmable gate arrays (FPGA) integrated circuits. This eliminates the need for high-performance cloud platforms, providing instant response at training, recognition and prediction in complex tasks, such as forecasting real useful life of aircraft engines, control of mobile autonomous objects, etc.

In conclusion, we note that a cortical mini-column in neurophysiology is a group of simultaneously activated neurons with almost identical receptor fields [12]. Although it has been known since 1957 [13, 14] that the cortical neurons are arranged in columns, it is widely debated what the function of the mini-column is, and whether the column is simply an economical way of packing neurons. However, if the cerebral cortex is an indexing system, then the formalism (2) for inverse patterns can be used to describe at a higher level a group of simultaneously activated neurons with identical receptor fields, - the neurons that send their brunching axons to different directions. As such, it can be hypothesized that the cortical **mini-column** serves as a **multiplier of connections** of a single feature (receptor field) to multiple classes (signal receivers), increasing the depth of associations. If the mini-column contains, on average, 100 neurons, then in the same proportion this expander increases the number of connections. Presumably, 100 neurons in a column is a more reliable design than a single neuron with a giant axon. Multiple axons constitute a more flexible design that allows a signal to easily travel to various destinations. However, a further increase in the number of connections (neurons in a mini-column) would lead to an increase in the energy consumption and to a decreasing generalization capability.